\documentclass[sn-mathphys,iicol]{sn-jnl}
\usepackage{array}
\usepackage{colortbl}

\usepackage{microtype}
\usepackage{graphicx}
\usepackage{subfigure}
\usepackage{balance}
\usepackage{booktabs} 
\usepackage{multirow}
\usepackage{makecell}
\usepackage{adjustbox}
\usepackage{color}
\usepackage{graphicx}%
\usepackage{multirow}%
\usepackage{amsmath,amssymb,amsfonts}%
\usepackage{amsthm}%
\usepackage{mathrsfs}%
\usepackage[title]{appendix}%
\usepackage{xcolor}%
\usepackage{textcomp}%
\usepackage{manyfoot}%
\usepackage{booktabs}%
\usepackage{algorithm}%
\usepackage{algorithmic}
\usepackage{listings}%
\usepackage{hyperref}
\usepackage{bbding}
\usepackage{tabularx}
\usepackage{url}
\usepackage{yfonts}
\usepackage{amssymb}
\usepackage{amsbsy}
\usepackage{amsmath}
\hypersetup{
    bookmarks=true,         
    unicode=false,          
    pdftoolbar=true,        
    pdfmenubar=true,        
    pdffitwindow=false,     
    pdftitle={Certificate},    
    pdfauthor={Dr. Harish Kumar},     
    pdfsubject={TEQIP certificates},   
    pdfcreator={Dr. Harish Kumar},   
    pdfproducer={},  
    pdfkeywords={Certificates,} {TEQIP} {Participation}, 
    pdfnewwindow=true,      
    colorlinks=false,       
    linkcolor=red,          
    citecolor=green,        
    filecolor=magenta,      
    urlcolor=cyan           
}

\theoremstyle{thmstyleone}%
\theoremstyle{thmstyletwo}%

\theoremstyle{thmstylethree}%

\begin{document}
\title[Enhancing Zero-Shot Image Recognition in Vision-Language Models through Human-like Concept Guidance]{Enhancing Zero-Shot Image Recognition in Vision-Language Models through Human-like Concept Guidance}


\author[1]{\fnm{Hui} \sur{Liu}}\email{liuhui3-c@my.cityu.edu.hk}
\author[2]{\fnm{Wenya} \sur{Wang}}\email{wangwy@ntu.edu.sg}
\author[3]{\fnm{Kecheng} \sur{Chen}}\email{ck.ee@my.cityu.edu.hk}
\author[1]{\fnm{Jie} \sur{Liu}}\email{jliu.ee@my.cityu.edu.hk}
\author[3]{\fnm{Yibing} \sur{Liu}}\email{lyibing112@gmail.com}
\author[1]{\fnm{Tiexin} \sur{Qin}}\email{tiexinqin@gmail.com}
\author[4]{\fnm{Peisong} \sur{He}}\email{gokeyhps@scu.edu.cn}
\author[5]{\fnm{Xinghao} \sur{Jiang}}\email{xhjiang@sjtu.edu.cn}
\author*[1]{\fnm{Haoliang} \sur{Li}}\email{haoliang.li@cityu.edu.hk}


\affil[1]{\orgdiv{Department of Electrical Engineering}, \orgname{City University of Hong Kong}, \orgaddress{ \state{Hong Kong SAR}}}
\affil[2]{\orgdiv{College of Computing and Data Science}, \orgname{Nanyang Technological University}, \orgaddress{\state{Singapore}}}
\affil[3]{\orgdiv{Department of Computer Science}, \orgname{City University of Hong Kong}, \orgaddress{\state{Hong Kong SAR}}}
\affil[4]{\orgdiv{School of cyber science and engineering}, \orgname{Sichuan University}, \orgaddress{\city{Chengdu}, \postcode{610065}, \country{China}}}
\affil[5]{\orgdiv{School of Electronic Information and Electrical Engineering}, \orgname{Shanghai Jiao Tong University}, \orgaddress{\city{Shanghai}, \postcode{200240}, \country{China}}}


\abstract{In zero-shot image recognition tasks, humans demonstrate remarkable flexibility in classifying unseen categories by composing known simpler concepts. However, existing vision-language models (VLMs), despite achieving significant progress through large-scale natural language supervision, often underperform in real-world applications because of sub-optimal prompt engineering and the inability to adapt effectively to target classes. To address these issues, we propose a Concept-guided Human-like Bayesian Reasoning (CHBR) framework. Grounded in Bayes' theorem, CHBR models the concept used in human image recognition as latent variables and formulates this task by summing across potential concepts, weighted by a prior distribution and a likelihood function. To tackle the intractable computation over an infinite concept space, we introduce an importance sampling algorithm that iteratively prompts large language models (LLMs) to generate discriminative concepts, emphasizing inter-class differences. We further propose three heuristic approaches involving Average Likelihood, Confidence Likelihood, and Test Time Augmentation (TTA) Likelihood, which dynamically refine the combination of concepts based on the test image. Extensive evaluations across fifteen datasets demonstrate that CHBR consistently outperforms existing state-of-the-art zero-shot generalization methods. }


\maketitle

\section{Introduction}
\label{sec:intro}
\begin{figure}[tb]
    \centering
        \centering
\includegraphics[width=\linewidth]{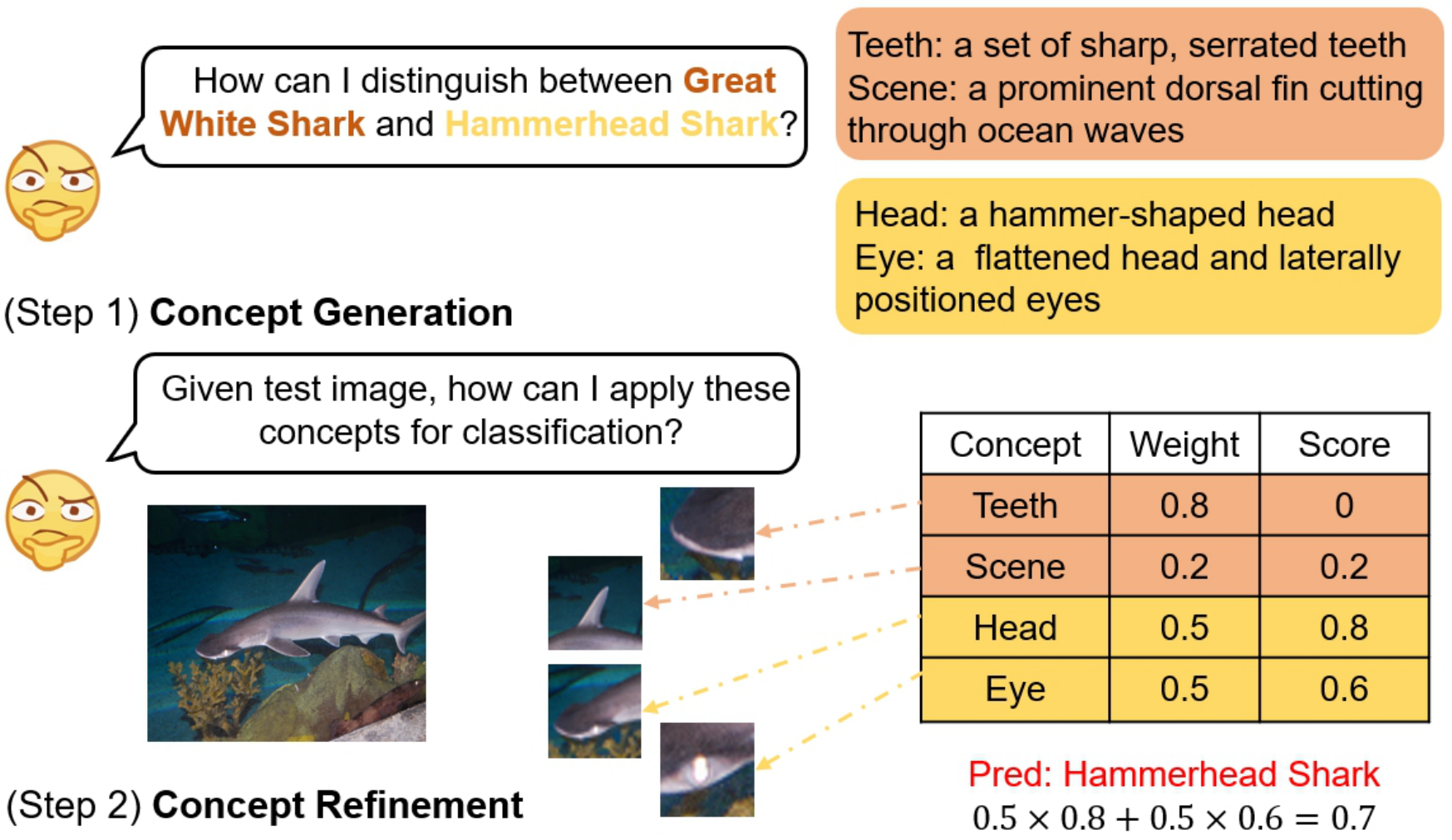}
    \caption{The pipeline of the concept-based human-like reasoning process for zero-shot image recognition. When the image depicts a shark on the ocean floor, the weight assigned to the concept scene is reduced, as humans can recognize that sharks are not exclusively on the sea but can also be found in deeper ocean environments.} 
    \label{fig:1}
\end{figure}

Humans, even as infants, exhibit remarkable flexibility in zero-shot image recognition, referring to classifying images into one of several classes that have not been seen before~\citep{xu2007word, lake2015human}.  This ability has been theoretically explained by the well-established ``recognition by components" model~\citep{biederman1987recognition}, which posits that humans can compose previously learned simpler concepts via reasoning to facilitate generalization on more complex tasks. As depicted in Fig.~\ref{fig:1},  when distinguishing between ``Great White Shark" and ``Hammerhead Shark", humans intuitively decompose these objects into basic components (e.g., ``teeth" and ``scene"). However, conventional image classification models trained on fixed-category object recognition datasets lack this compositional generalization ability~\citep{mishra2022learning, chen2024progressive} as they cannot spontaneously combine known features to infer new classes, thus limiting their capacity for zero-shot image classification. 

Vision-language models (VLMs) have revolutionized this area by leveraging large-scale natural language supervision~\citep{clip, blip2, evaclip, llava}, enabling the comprehension of a wider range of visual concepts far beyond traditional models.  A prominent example is CLIP~\citep{clip}, which learns to associate images with their corresponding captions from massive datasets of image-text pairs. This enables CLIP to perform zero-shot image recognition without task-specific fine-tuning by using text descriptions to reference visual concepts\footnote{Since CLIP is trained on hundreds of millions of text-image pairs sourced from the internet, ~\citet{clip} evaluate its zero-shot capability at the dataset level.} Specifically, class labels are incorporated into a predefined prompt template (e.g., ``A photo of \{class\}.”), which is then processed by CLIP’s text encoder. The test image is projected into the same embedding space using the visual encoder, and the class with the highest similarity between the image and text embeddings is selected as the final prediction.

Despite the promise of such prompt-based methods, they often yield unsatisfactory results in real-world applications potentially because of sub-optimal prompt engineering~\citep{coop, cocoop} (e.g., ``A photo of a \{class\}." may not be effective when the class is not the main subject of the image) and inadequate adaption to target classes~\citep{clipadapter,tipadapter, yu2023task}. To alleviate this issue, existing approaches~\citep{shu2022tpt, ma2024swapprompt, feng2023diverse, matcvpr24} mainly employ test time augmentation techniques~\citep{kimura2021understanding,wang2020tent, liang2024comprehensive}. However, these methods do not fully exploit the rich visual knowledge captured by VLMs, encompassing a broad range of visual concepts from low-level features (e.g., textures) to high-level abstractions (e.g., scenes). Additionally, some recent works~\citep{pratt2023does, zeroshotcvpr23, novack2023chilszeroshot} expand class definitions by class-specific descriptions, which, however, often struggle in fine-grained classification tasks where only subtle differences can distinguish the target classes.  For instance, for ``Great White Shark" and ``Hammerhead Shark" in Fig. \ref{fig:1}, a generic depiction like ``a dorsal fin" is less informative as both kinds of sharks have this feature, while ``a hammer-shaped head" would provide a clearer distinction. 

To tackle these challenges, and inspired by the human cognition process, we propose a novel \textbf{C}oncept-guided \textbf{H}uman-like \textbf{B}ayesian \textbf{R}easoning (CHBR) framework. This framework models the concept used in human image recognition as latent variables and formulates this task by summing across potential concepts, weighted by a prior distribution and a likelihood function. Our proposed framework focuses on two critical issues of imitating human-like reasoning. \textbf{First}, humans can acquire useful concept priors through environmental feedback \citep{kahneman2011thinking} or instruction from experienced individuals \citep{hinton2015distilling}, whereas the automatic discovery of concepts is an unresolved issue for models. This challenge arises because the space of possible concepts is effectively infinite, and performing inference over such a vast concept space incurs significant computational costs \citep{zhang2024human, lake2015human, liu2023unsupervised, ellis2024human}.  To address the computational intractability of reasoning over this infinite concept space, we introduce a Monte Carlo importance sampling algorithm that iteratively prompts LLMs to generate discriminative textual concepts for each class. Crucially, these concepts highlight inter-class differences instead of merely describing individual classes, thereby enriching the sampled concept space and sharpening classification boundaries. Each concept's prior distribution is then derived directly from this sampling process. \textbf{Second}, while humans can refine their knowledge through slow, deliberate reasoning \citep{kahneman2011thinking}, even advanced VLMs often struggle to adapt flexibly during test time, as they are typically data-hungry and require large amounts of data to learn successfully. Additionally, LLMs may be susceptible to hallucinations and intrinsic biases, which can cause the generated concepts to be misaligned with the test image. For instance, a ``head" concept may be occluded by other objects in the image, rendering it ineffective for the given test case in Fig. \ref{fig:1}.  To address these limitations and to adaptively model the likelihood associated with each concept, we propose three approaches: \textit{Average Likelihood}, \textit{Confidence Likelihood}, and  \textit{TTA Likelihood} to refine the importance of each concept based on the specific test image, enabling CHBR to adapt dynamically and generalize more effectively during inference. 

To evaluate the effectiveness of our framework, we conduct experiments across fifteen image recognition tasks. The results demonstrate that CHBR consistently outperforms existing state-of-the-art methods. We summarize our main contributions as follows:
\begin{itemize}
    \item We propose a novel Concept-guided Human-like Bayesian Reasoning (CHBR) framework for CLIP-like VLMs-based zero-shot image recognition, which operates by marginalizing over a space of potential concepts, with each concept weighted by a prior distribution and a likelihood function
    
    \item We introduce an importance sampling technique to approximate the concept space, along with three heuristic implementations of the likelihood function,  allowing users to select the best approach suited to their requirements. Especially, Averaging Likelihood and Confidence Likelihood are particularly effective in fine-grained image classification tasks, while TTA likelihood demonstrates robustness across both ImageNet and fine-grained scenarios. However, the inference time of Averaging Likelihood and Confidence Likelihood is comparable to  directly prompting CLIP for classification, achieving significantly lower latency than TTA Likelihood and other test-time adaptation baselines.
    
    \item We demonstrate the effectiveness of CHBR across diverse image recognition tasks, highlighting its potential for future zero-shot image recognition challenges.
\end{itemize}

\section{Related Works}
\label{sec:related}
\subsection{Vision-language model generalization} 
Vision language model generalization has garnered substantial attention in recent years due to the inherent difficulty of fine-tuning all parameters while maintaining robust generalization capabilities, given the extreme scarcity of training examples per class~\citep{coop, lester2021power}. Broadly, these techniques can be classified into two categories based on the availability of a few labeled examples for each class: few-shot methods and zero-shot methods. 

In the few-shot scenario, mainstream approaches often utilize parameter-efficient fine-tuning techniques, such as soft prompt tuning~\citep{coop, cocoop, zhu2023prompt, chen2022plot, huang2022unsupervised, wang2023improving} and adapter-based methods~\citep{tipadapter, clipadapter}, to introduce additional trainable parameters into a frozen VLM. For example, CoOp~\citep{coop} inserts learnable continuous tokens into the original prompt template and updates these token embeddings by minimizing the classification loss, thereby overcoming the need for manual prompt engineering. Another line of research follows a learning-to-compare paradigm~\citep{vinyals2016matching}, which generates prototypes for each class based on support images and directly compares test images with these prototypes for classification~\citep{hou2022closer, martin2024transductive, zanellaboostingnips24, silva2024closer, udandarao2023sus, zhang2023prompt}. 

In the zero-shot setting, one group of works employs data augmentation or diffusion models to produce diverse views for test images and encourage consistent predictions across these views by prompt tuning~\citep{shu2022tpt, ma2024swapprompt, feng2023diverse} or view-specific weight optimization~\citep{matcvpr24}. These methods can be integrated into our framework as the test time augmentation likelihood. Another group leverages LLMs to generate more detailed prompts that describe the test class~\citep{pratt2023does}, or use hierarchical word structures (e.g., subclass belong to the test class)~\citep{zeroshotcvpr23, novack2023chilszeroshot} to enrich class names. However, these methods struggle to generate discriminative concepts for complex real-world situations or are constrained by the need for datasets with well-defined semantic structures. Moreover, these approaches only consider the concept prior without adjusting the importance of different concepts, as human beings do. In contrast, our framework integrates both concepts prior and likelihood, allowing for more adaptive and discriminative zero-shot image recognition. Additionally, another line of work focuses on transforming images into concepts inside predefined concept banks and mapping these concepts to target classes~\citep{shang2024incremental}. However, such concept banks are static and cannot be updated in real time to adapt to specific image recognition tasks. In contrast, our method leverages LLM-derived concept priors learned from extensive real-world data, enabling more flexible image recognition.

\subsection{Concept discovery} 
Concept discovery refers to inferring general patterns from specific examples, widely utilized across cognitive science and artificial intelligences~\citep{ellis2021dreamcoder, liu2023unsupervised, chater2008probabilistic}. A principal approach to this task is first to define a hypothesis space of potential concepts and then use Bayesian inference to determine which hypothesis most likely generates the observed data probabilistically~\citep{ellis2021dreamcoder, saad2019bayesian, tian2020learning}. This paradigm raises the issue of balancing the expressiveness of the hypothesis space with computational feasibility, as the number of possible concept combinations grows exponentially with the increase of the size of the hypothesis space. A common strategy to address this challenge is to sample representative concepts to approximate the full hypothesis space~\citep{ellis2024human} or mine latent concepts from data and assign meaning to them using supervised learning~\citep{glanois2022neuro}. In our setting, given a set of image classes, we utilize LLMs to iteratively sample concepts that effectively discriminate these classes, aiming to approximate the concept prior that humans might use in the same classification task. Recent work~\citep{elicitpriorzhu} has validated that priors elicited from LLMs based on Monte Carlo sampling qualitatively align with human-like priors because of the extensive pretraining on vast amounts of real-world data. 

\begin{figure}[tb]
    \centering
        \centering
\includegraphics[width=\linewidth]{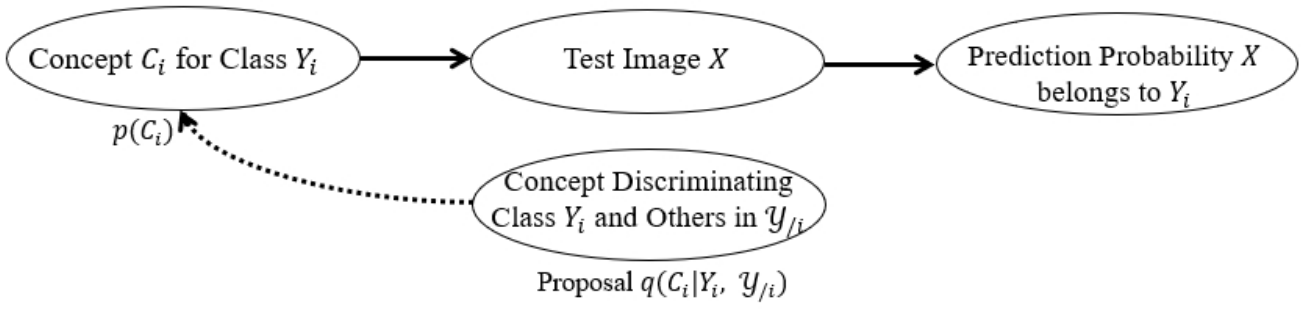}
    \caption{Bayesian network that conceptualizes human-like zero-shot image recognition process. } 
    \label{fig:2}
\end{figure}
\section{Preliminary}
\label{sec:preliminary}
In a typical zero-shot image recognition task, let us consider a set of candidate image classes $\mathcal{Y} = \{ Y_1, \dots, Y_K \}$, where $Y_i$ denotes the $i$-th image class and $K$ represents the total number of classes. Given a test image $X$, for CLIP-like VLMs, the $i$-th class is represented as ``A photo of a $Y_i$.". This text description and the test image $X$ are encoded by CLIP's text and image encoders to produce the corresponding normalized textual and visual embeddings. Subsequently, the probability that $X$ belongs to $Y_i$, denoted $p(Y_i |X)$, can be computed as the cosine similarity between the text and image embeddings. The predicted class label is then determined by selecting the class with the highest probability.
\section{Methodology}
\label{sec:method}
Humans possess an exceptional ability for zero-shot image recognition, primarily due to their extensive prior knowledge of the world and capability to efficiently refine this knowledge to adapt to unseen inputs rapidly. Drawing inspiration from this cognition process, we propose a Concept-based Human-like Bayesian Reasoning (CHBR) framework.  In this section,  we first conceptualize image classification as a summation of potential concepts over a latent space, where each concept is weighted by a prior distribution and a likelihood and seek an approximation of this expectation using a proposal distribution because of the infinite nature of the concept space in Sec.~\ref{sec:4.1}. The corresponding probabilistic graphical model is depicted in Fig. \ref{fig:2}. Then, we detail how to obtain the proposal distribution and likelihood function in Sec. \ref{sec:4.2} and Sec. \ref{sec:4.3}, respectively. The outlined algorithm pipeline is described in Algorithm \ref{al:1}.
\subsection{Concept-based Bayesian reasoning framework}
\label{sec:4.1}
While the naive zero-shot image recognition paradigm in Sec.~\ref{sec:preliminary} shows empirical efficacy, it relies solely on semantic information extracted by CLIP's encoders, neglecting potentially valuable contextual clues (e.g., relations between classes in $\mathcal{Y}$) and prior knowledge that humans might use for zero-shot image classification. Therefore, we introduce latent concepts that characterize each class from the perspective of human cognition. Specifically, we hypothesize that the test image $X$ can be generated based on a set of latent concepts, denoted as $\mathcal{C} = \{ C_1, \ldots, C_K \}$, where each variable $C_i$ represents all possible concepts corresponding to class $Y_i$ and the size of the sampling space of  $C_i$ is infinite. Then, the probability  $p(Y_i|X)$, given latent concepts $C_i$, can be written as:
\begin{equation}
    p(Y_i|X) = \sum_{C_i} p(Y_i|X, C_i) p(C_i|X),
\end{equation}where $p(C_i|X)$ represents the posterior probability over the latent concept $C_i$ given the image $X$. Since latent concepts are expressed in natural language, they can be seamlessly incorporated into the original prompt ``A photo of a $Y_i$.” to form a more discriminative prompt, such as ``A photo of a $Y_i$ with $C_i$.”. Then, $p(Y_i|X, C_i)$ can be computed using the same approach as the typical zero-shot setting. 
However, directly estimating $p(C_i|X)$ is computationally infeasible due to the large space of possible concepts and the unaffordable cost of generating concepts in real-time for every test image. As a result, we employ Bayes' theorem to reformulate $p(C_i|X)$ as:
\begin{equation}
    p(C_i|X) = \frac{p(X|C_i) p(C_i)}{p(X)}\propto p(X|C_i) p(C_i), 
\end{equation}where $p(C_i)$ represents the prior probability distribution of $C_i$. It serves as a preliminary understanding of the potential concept describing the image class $Y_i$, thus encoding the instinctive world knowledge of humans for image recognition~\citep{kahneman2011thinking}. The likelihood term $p(X|C_i)$ models the compatibility between the image $X$ and the concept $C_i$, refining the posterior toward reasonable generalizations for test-time adaptation. $p(X)$ is omitted because it is a constant that does not affect the relative magnitude of $p(C_i|X)$. Finally, the classification decision is made by marginalizing over $C_i$:
\begin{equation}
\label{eq:3}
    p(Y_i|X) \approx \sum_{C_i} p(Y_i|X, C_i)p(X|C_i) p(C_i),
\end{equation}which allows for integrating more granular information than the naive zero-shot classification method that solely computes $p(Y_i |X)$ via the cosine similarity between the embeddings of enhanced class description and image. 

Although the method for obtaining $p(Y_i|X, C_i)$ has been established, determining how to obtain the concept $C_i$ is still agnostic. One of the central challenges in concept discovery is defining an expressive but computationally feasible concept space, given the effectively infinite set of possible concepts. To address this issue, we draw inspiration from importance sampling theory~\citep{kloek1978bayesian} to introduce a proposal distribution, denoted as $q(C_i)$, from which concept samples can be efficiently generated. Then we draw a relatively modest number of concept samples (on the order of tens to hundreds), denoted as $\{C_{i,1}, ..., C_{i,M_i}\}$ for class $Y_i$ in an IID manner from $q(C_{i})$. This approach reduces the intractable sum over an infinite set of concepts in Eq.~\ref{eq:3} to a finite sum over the sampled concepts as:
\begin{equation}
\label{eq:6}
        p(Y_i|X) \approx \sum_{C_{i,j}}^{M_i} p(Y_i|X, C_{i,j})p(X|C_{i,j}) w_{i,j},
\end{equation}where the importance weight $w_{i,j} = \frac{p(C_{i,j})}{q(C_{i,j})}$ corrects the bias caused by sampling from the proposal distribution  $q(C_i)$. At this point, we have completed the concept-based Bayesian reasoning framework theoretically, and we will detail how to implement the suitable proposal distribution $q(C_i)$ and concept likelihood $p(X|C_{i,j})$ in the following.
\subsection{Sampling Strategy for concept prior}
\label{sec:4.2}
\begin{algorithm}
\caption{Pipeline of our CHBR framework}\label{alg:chbr}
\label{al:1}
\begin{algorithmic}[1]
\STATE \textbf{Input:} The image classes set $\mathcal{Y}$ and  test image $X$ for an image recognition task
\STATE \textbf{Output:} Predicted class $\hat{Y}$ for Test image $X$ 
\STATE \textbf{Initialization:} The type of the likelihood function $p(X|C)$ (Average Likelihood, Confidence Likelihood or TTA Likelihood)
\FOR{each class $Y_i \in \mathcal{Y}$}
   \STATE  Sample $M_i$ concepts $\{C_{i,1}, \dots, C_{i,M_i}\}$ from proposal distribution $q(C_{i}|\mathcal{Y}_{/i}, Y_i)$ based on sampling strategy in Sec. \ref{sec:4.2}
    \STATE Compute each importance weight $w_{i,j}$ using Eq. \ref{eq:weight} for each $C_{i,j}$ where $j \in [1, M_i]$
    \STATE Compute each $p(C_{i,j}|X)$  where $j \in [1, M_i]$ using initialized likelihood function described in Sec. \ref{sec:4.3}
    \STATE Compute $p(Y_i|X)$ using Eq. \ref{eq:6} 
\ENDFOR
\STATE \textbf{Final Decision:} Select the class with the highest posterior probability: $\hat{Y} = \arg\max_{Y_i} p(Y_i|X)$
\end{algorithmic}
\end{algorithm}
When devising a suitable proposal distribution $q(C_i)$ where the sampled concepts can occupy a significant portion of the mass of $p(C_i)$, it would be helpful to consider the nature of this prior in image recognition tasks. Intuitively, a robust concept space should describe a class accurately and enhance discriminability between classes. Therefore, an ideal distribution of concept $C_i$ should be conditional on both the target class $Y_i$ and its complement $\mathcal{Y}_{/i}$ (i.e., all classes except $Y_i$). As such, $q(C_{i} |\mathcal{Y}_{/i}, Y_i)$ can become an optimal proposal distribution $q(C_i)$ for approximating $p(C_i)$. To implement Monte Carlo importance sampling from $q(C_{i}| \mathcal{Y}_{/i}, Y_i)$, we propose a large language model driven sampler as follows: For each class $Y_i$, we randomly draw $H$ candidate classes from the image class set $\mathcal{Y}$, forming a candidate set $L_{i,j}$ for $Y_i$. The LLM is then prompted to generate candidate concepts $C_{i,j}$ to describe $Y_i$ in a way that maximally distinguishes it from the other classes in  $L_{i,j}$. To assess the discriminability of $C_{i,j}$, we use the distractor classes in $L_{i,j}$ to create a discriminative test, deciding whether LLM can accurately predict $Y_i$ from these distractor ones given $C_{i,j}$. The success rate $s_{i,j}$ regarding that $C_{i,j}$ passing this test is recorded.  Given that the true prior distribution $p(C_{i,j})$ is typically difficult to obtain, we approximate it using the success rate $s_{i,j}$. Finally, importance weight $w_{i,j}$ can be computed as:
\begin{equation}
\label{eq:weight}
w_{i,j} = \frac{s_{i,j}}{q(C_{i,j}|\mathcal{Y}_{/i}, Y_i)},
\end{equation}where $q(C_{i,j}|\mathcal{Y}_{/i}, Y_i)$ equals to $1$ for simplification purpose following recent practice~\citep{ellis2024human}. Due to the sum operation in Eq. \ref{eq:6}, those concepts with more semantically similar ones in the sampling space will have a larger influence on the final classification decision. Details of the whole sampling algorithm and the prompts to interact with LLM are provided in Appendix \ref{app:1}.
\subsection{Instantiation for concept likelihood}
\label{sec:4.3}
After completing the concept sampling process mentioned above, we seek an estimation of the likelihood function $p(X|C_{i,j})$, which can refine the importance of each concept based on the test image $X$ to enable test time adaptation for our framework. We propose three implementations for this estimation: Average Likelihood, Confidence Likelihood, and Test-Time Augmentation Likelihood (TTA Likelihood).

\noindent\textbf{Average Likelihood}: This approach assumes that each concept contributes equally to the recognition of the input $X$. Consequently, we implement $p(X|C_{i,j})$ using uniform weighting, serving as a baseline where no particular concept is favored over others.

\noindent\textbf{Confidence Likelihood}: While the sampled concepts are designed to be discriminative for classifying existing classes from a human-like perspective, some may fail to respond strongly to images from their corresponding classes. This could be caused by factors such as occlusion, variations in camera angles, or the model's limited capacity to capture certain nuances. To address this issue, we propose Confidence Likelihood, which prioritizes concepts that exhibit higher similarity to the given image than others within the same class. Formally, the confidence-based likelihood for concept $C_{i,j}$ and test image $X$ is computed as:
\begin{equation}
p(X|C_{i,j}) = \frac{\exp(p(Y_i | X, C_{i,j}) \cdot \tau)}{\sum_i \exp(p(Y_i | X, C_{i,j}) \cdot \tau)},
\end{equation}where $\tau$ is a temperature parameter that controls the sharpness of the distribution, and $p(Y_i | X, C_{i,j})$ represents the similarity between the text ``A photo of a $Y_i$ with $C_{i,j}$." and the image $X$. This formulation enables the likelihood function to focus on the concepts most relevant to the input image, thereby enhancing discriminative power.

\noindent\textbf{TTA Likelihood:} Inspired by the powerful test-time augmentation (TTA) techniques~\citep{wang2020tent}, we introduce TTA Likelihood that maximizes the consistency among multiple augmented views of the input image $X$. Specifically, we generate $N$ augmented views, denoted as $\{X_1, X_2, \dots, X_N\}$ for $X$. For each class $Y_i$, we introduce a learnable parameter $\theta_i \in \mathbb{R}^{M_i}$, which is normalized via a softmax operation to obtain the normalized likelihood $p(X_n|C_{i,j})$ for each concept $C_{i,j}$, denoted as $\hat{\theta}_{i,j}$ where $\sum_j\hat{\theta}_{i,j}=1$.
The classification probability for each view $X_n$ can then be computed as:
\begin{equation}
p(Y_i|X_n) \approx \sum_{j=1}^{M_i} p(Y_i| X_n, C_{i,j}) \hat{\theta}_{i,j} w_{i,j}.
\end{equation}The overall prediction probability for the original image $X$ is then obtained by averaging over all augmented views as $\hat{p}(Y_i|X) = \frac{1}{N} \sum_{n=1}^{N} p(Y_i|X_n)
$. Finally, we minimize the entropy of this averaged prediction probability distribution to encourage consistency across all augmented views:
\begin{equation}
\arg\min_{\theta} -\sum_{i=1}^{K} \hat{p}(Y_i| X) \log \hat{p}(Y_i| X),
\end{equation}where $K$ is the number of candidate image classes. In addition, to mitigate the noise introduced by random augmentations, we leverage a confidence selection mechanism following~\citet{shu2022tpt}. Specifically, we filter out views that produce high-entropy predictions, as these views may lack critical information for correct classification. For example, a random crop may exclude essential parts of the image. We retain only the views that rank in the top-$r$ percent of the total number of views regarding their entropy to ensure that the final prediction is based on the most informative views.

\section{Experiments}
\label{sec:exres}
\subsection{Datasets} 
\begin{table*}[h]
\centering
\caption{\centering Statistics of fifteen datasets used in our work.}
\label{tab:datasets_details}
\begin{adjustbox}{width=0.7\linewidth, center}
    \begin{tabular}{l@{\hskip 1em}c@{\hskip 1em}c}
    \toprule
        \textbf{Dataset} & \textbf{Number of Classes} & \textbf{Test Set Size} \\
    \midrule
    SUN397~\citep{sun397} & 397 & 19,850 \\
    Aircraft~\citep{maji2013fine} & 100 & 3,333 \\
    EuroSAT~\citep{helber2019eurosat} & 10 & 8,100 \\
    StanfordCars~\citep{krause20133d} & 196 & 8,041 \\
    Food101~\citep{food101} & 101 & 30,300 \\
    OxfordPets~\citep{parkhi2012cats} & 37 & 3,669 \\
    Flower102~\citep{nilsback2008automated} & 102 & 2,463 \\
    Caltech101~\citep{fei2004learning} & 100 & 2,465 \\
    DTD~\citep{cimpoi2014describing} & 47 & 1,692 \\
    UCF101~\citep{soomro2012ucf101} & 101 & 3,783 \\
    \midrule
    ImageNet~\citep{deng2009imagenet} & 1,000 & 50,000 \\
    ImageNet-A~\citep{imageneta} & 200 & 7,500 \\
    ImageNet-V2~\citep{imagenetv2} & 1,000 & 10,000 \\
    ImageNet-R~\citep{imagenetr} & 200 & 30,000 \\
    ImageNet-Sketch~\citep{imagenetsketch} & 1,000 & 50,889 \\
    \bottomrule
    \end{tabular}
\end{adjustbox}
\end{table*}

Following the evaluation protocol of prior works~\citep{matcvpr24, shu2022tpt, coop}, we primarily assess our framework on ten fine-grained classification datasets, including automobiles (Cars~\citep{krause20133d}), textures (DTD~\citep{cimpoi2014describing}), human actions (UCF101~\citep{soomro2012ucf101}), aircraft types (Aircraft~\citep{maji2013fine}), satellite imagery (EuroSAT~\citep{helber2019eurosat}), pet breeds (Pets~\citep{parkhi2012cats}), flowers (Flower102~\citep{nilsback2008automated}), food items (Food101~\citep{food101}), scenes (SUN397~\citep{sun397}), and general objects (Caltech101~\citep{fei2004learning}). We additionally evaluate the robustness of our framework to natural distribution shifts on four ImageNet variants, involving ImageNet-A~\citep{imageneta}, ImageNet-V2~\citep{imagenetv2}, ImageNet-R~\citep{imagenetr}, ImageNet-Sketch~\citep{imagenetsketch}, as out-of-distribution (OOD) data for ImageNet~\citep{deng2009imagenet}. The statistics of all datasets are shown in Table \ref{tab:datasets_details}.

\subsection{Baselines} 
In line with previous study~\citep{matcvpr24}, we compare our framework with four widely recognized zero-shot image recognition baselines, involving \textit{CLIP}~\citep{clip}, CLIP + E~\citep{ 
clip}, TPT~\citep{shu2022tpt} and MTA~\citep{matcvpr24}. These baselines differ in their choice of prompt templates and their use of test time augmentations. \textit{CLIP} and CLIP + E are two versions adopted by~\citet{clip} while the former uses a default prompt ``A photo of a \{class\}." and the latter uses the ensemble of 80 hand-crafted prompts. Built upon \textit{CLIP},  TPT and MTA generate multiple augmented views of the test image to enhance prediction robustness. TPT minimizes the entropy of the prediction distributions among all views via prompt tuning. MTA, on the other hand, optimizes an importance score for each view to find the mode of the density of all views and enforces similar importance scores for views with similar prediction distributions. We evaluate three likelihood-based variants for our proposed concept-based human-like Bayesian Reasoning (CHBR) framework: Average Likelihood, Confidence Likelihood and TTA Likelihood. 
\begin{table*}[t!]
\caption{Performance of zero-shot methods on ten fine-grained classification datasets. We highlight the best and second best results by {\bf bolding} and \underline{underlining} them, respectively. The error bar of three likelihood functions is reported in Table \ref{tab:zero_shot_datasets_bar} in the Appendix.}
\label{tab:zero_shot_datasets}
\centering
\begin{adjustbox}{width=\linewidth, center}
\begin{tabular}{lccccccccccc}
\toprule
Method & SUN397 & Aircraft & EuroSAT & Cars & Food101 & Pets &  Flower102 & Caltech101 & DTD & UCF101 & Average
\\ \midrule 
CLIP~\citep{clip} & 62.32 & 23.91 & 42.22 & 65.51 & 82.21 & 87.27 & 67.40 & 92.21 & 44.39 & 64.26 & 63.17 \\
CLIP + E.~\citep{clip} & 65.13 & 23.67 & 47.73 & 66.27 & 82.28 & 87.11 & 66.10 & 93.06 & 44.04 & 65.00 & 64.13\\
\midrule
TPT~\citep{shu2022tpt} & 65.41 & 23.10 & 42.93 & 66.36 & \underline{84.63} & 87.22 & 68.86 & \underline{94.12} & 46.99 & 68.00 & 64.76\\
MTA~\citep{matcvpr24}  & 64.98 & 25.32 & 38.71 & \underline{68.05} & \bf{84.95} & 88.22 & 68.26 & \bf{94.13} & 45.59 & 68.11 & 64.63 \\
\midrule
CHBR w/ Averaging   Likelihood & 65.54 & 25.74 & \bf{48.96} & 66.51  & 82.46 & 88.14 & 68.41 & 93.91 & 50.65 & 69.13 & 65.95 \\
CHBR w/  Confidence   Likelihood & \underline{66.25} & \underline{26.28} & \underline{48.89} & 66.62 & 82.62 & \bf{90.08}  & \bf{71.30} & 93.59 & \underline{51.95}  & \bf{69.57}  & \bf{66.72} \\
CHBR w/  TTA Likelihood  & \bf{66.89} & \bf{28.20} & 41.69 & \bf{68.86} & 82.68 & \underline{89.89} & \underline{70.40} & 92.74 & \bf{52.90} & \underline{69.39} & \underline{66.34} \\
\bottomrule
\end{tabular}
\end{adjustbox}
\end{table*}

\begin{table*}[t!]
\caption{Performance of zero-shot methods on ImageNet datasets. We highlight the best and second best results by {\bf bolding} and \underline{underlining} them, respectively.} 
\label{tab:zero_shot_imagenet}
\centering
\begin{adjustbox}{width=0.9\linewidth, center}
\begin{tabular}{lccccccc}
\toprule 
\multicolumn{1}{l}{\bf Method}  
&\multicolumn{1}{c}{\makecell{\bf ImageNet}}
&\multicolumn{1}{c}{\makecell{\bf -A}} 
&\multicolumn{1}{c}{\makecell{\bf -V2}} 
&\multicolumn{1}{c}{\makecell{\bf -R}}
&\multicolumn{1}{c}{\makecell{\bf -Sketch}}
&\multicolumn{1}{c}{\bf Average} 
&\multicolumn{1}{c}{\bf OOD Average } 
\\ \midrule
CLIP~\citep{clip}   
     & 66.73  & 47.80 & 60.84 & 73.99 & 46.15 &  59.10    &  57.20  \\
CLIP + E .~\citep{clip}   
         & 68.38 & 49.95 & 61.91 & \textbf{77.71} & 48.26 & 61.23  &  59.46   \\
\midrule
TPT~\citep{shu2022tpt}         & 68.94 & 54.63 & 63.41 & \underline{77.04} & 47.97  & 62.40   &  60.76 \\
MTA~\citep{matcvpr24}      & \underline{69.29} & \underline{57.41} & \underline{63.61} & 76.92 & \textbf{48.58} & \underline{63.16}  &  \underline{61.63}  \\
\midrule 
CHBR w/  Averaging   Likelihood        & 67.42 & 47.84 & 61.00 & 74.59 &46.37& 59.44  & 57.45 \\
CHBR w/  Confidence   Likelihood     & 67.88 & 48.28 & 61.35 & 74.45 & 46.55 & 59.70  &  57.66 \\
CHBR w/  TTA Likelihood      & $\textbf{70.05}$ & \textbf{58.32} & \textbf{63.79} & 76.97 & \underline{48.47} & \textbf{63.52}   &  \bf{61.89} \\
\bottomrule
\end{tabular}
\end{adjustbox}
\end{table*}
\subsection{Implementation details}
To obtain the concept prior, we employ GPT-4o mini to construct an importance sampler. This model is a more cost-efficient LLM than GPT-4 Turbo\footnote{\url{https://platform.openai.com/docs/models/gpt-4-turbo}}. We set the size of sampling windows as 4 for ten fine-grained image classification datasets  ($H=4$)  and 9 for the ImageNet dataset  ($H=9$), aiming to enhance sampling efficiency given that ImageNet contains 1,000 classes. For all datasets, we perform 100 sampling iterations ($M_i=100$ for each class). The verification times for each concept in the discriminative test are set to 5 ($Z=5$). To obtain the likelihood function, the temperature parameter $\tau$ is fixed at 3 across all datasets for Confidence Likelihood. For TTA Likelihood, we follow TPT~\citep{shu2022tpt} to augment a single test image using random resized crops for ImageNet-series datasets and AugMix~\citep{hendrycksaugmix} for fine-grained classification datasets. This augmentation process produces a batch of 64 images  ($N=64$) involving the original test image, from which we select the top 10\% ($r=10$) most confident samples. Additionally, we optimize the learnable weight assigned to each concept via AdamW optimizer, using a learning rate of 1 and learning steps of 30. We adopt the default parameter settings for other baselines as reported in their respective studies. Unless otherwise specified, we use CLIP with the ViT-B/16 backbone as the primary vision-language model. To avoid randomness, we report the averaged top-1 accuracy of three different random seeds for the main experiments. The code will be made publicly available upon the acceptance of the paper.

\subsection{Zero-shot Performance Analysis}
\begin{figure}[htbp]
    \centering
        \centering
\includegraphics[width=0.85\linewidth]{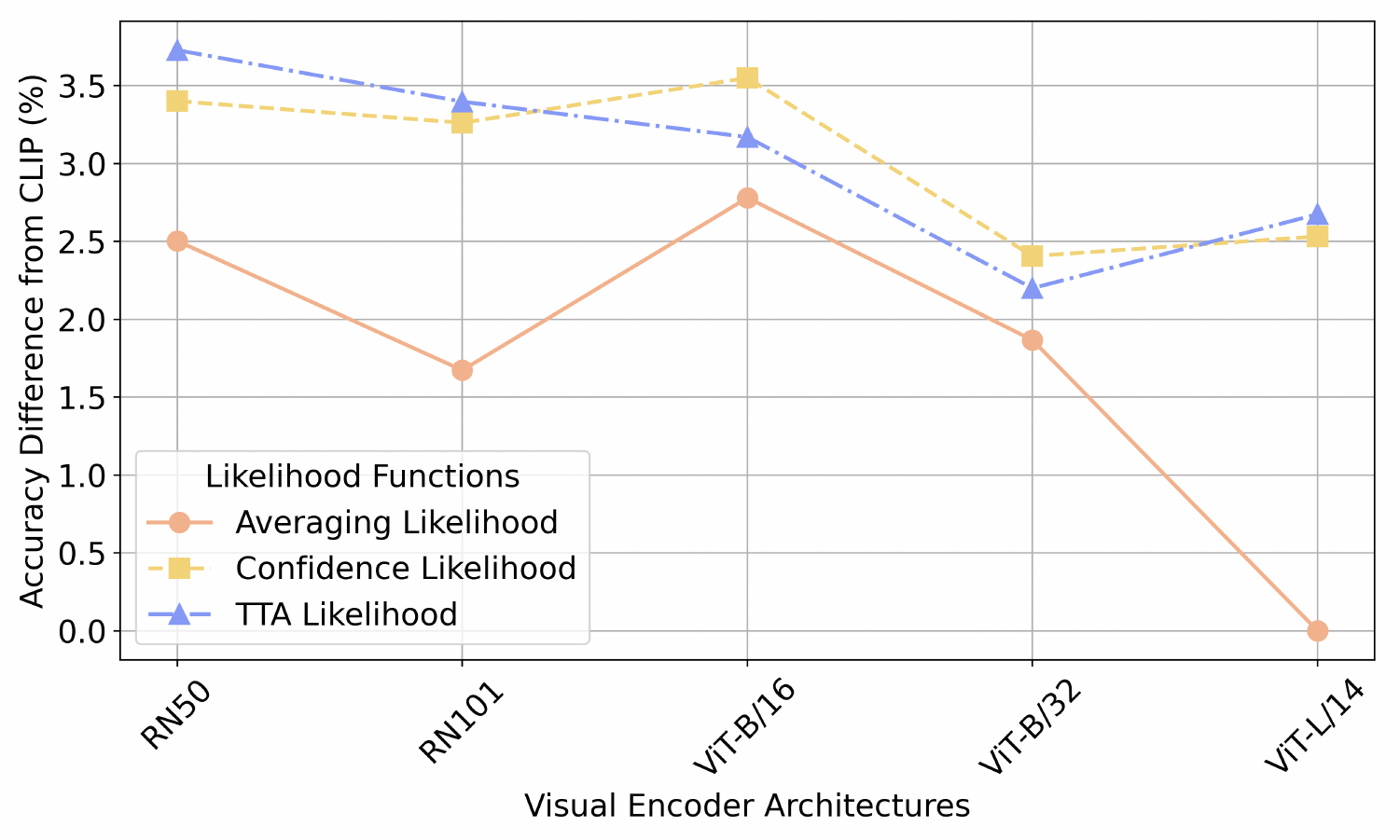}
    \caption{The difference in averaged top-1 accuracy between CHBR, inestantiated with various visual encoders and the baseline \textit{CLIP} across \textbf{ten} fine-grained image recognition tasks. Hyperparameters are kept consistent across all experiments.} 
    \label{fig:3}
\end{figure}
\textbf{CHBR exhibits significant advantages over existing zero-shot generalization methods in fine-grained image recognition tasks}. As shown in Table \ref{tab:zero_shot_datasets}, when instantiated with Averaging Likelihood, CHBR outperforms the current state-of-the-art method, TPT, by 1.19\% in terms of average accuracy across ten datasets. This improvement validates the effectiveness of the concepts generated through our proposed importance sampling algorithm, particularly in fine-grained image classification settings, where class distinctions are often visually subtle. Moreover, these results suggest that priors derived from the LLM based on textual class names have the potential to align with real-world priors of humans in image classification tasks. Notably, when Confidence Likelihood and TTA Likelihood are employed to filter out concepts that exhibit minimal response to test images or show inconsistent prediction distributions across different augmentation views, CHBR achieves consistent improvements over baselines across all datasets. These findings underscore the validity of concept refinement based on the characteristics of test images. Moreover, they highlight that different images may exhibit varying affinities for various concepts, demonstrating the adaptive nature of the zero-shot image recognition task.

\begin{figure}[tb]
    \centering
        \centering
\includegraphics[width=0.85\linewidth]{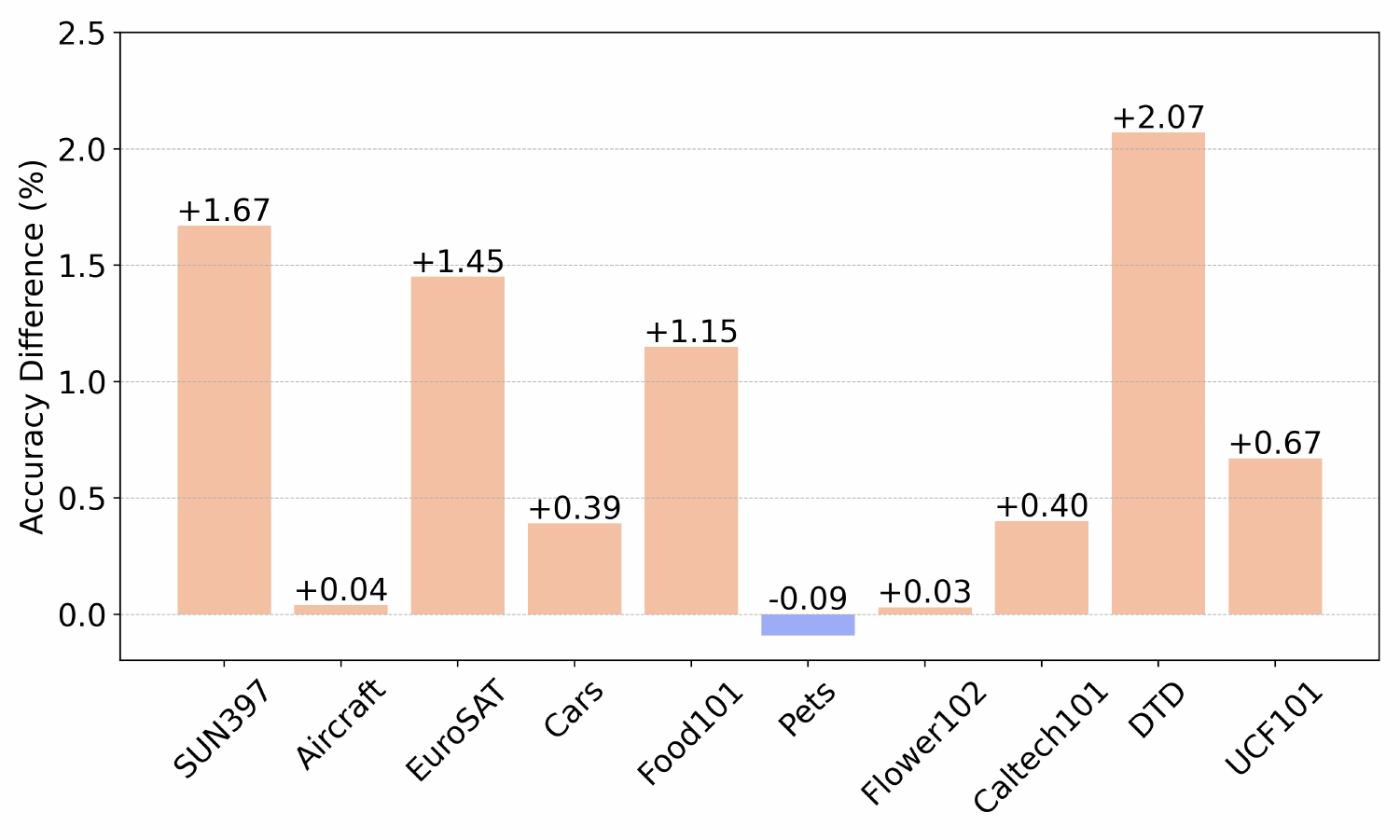}
    \caption{Top-1 Accuracy difference between discriminative and descriptive concepts on ten fine-grained datasets. We use Averaged Likelihood for illustration.} 
    \label{fig:4}
\end{figure}

\textbf{CHBR demonstrates robustness across various domain shifts}. As shown in Table \ref{tab:zero_shot_imagenet}, we transfer the concept generated from the ImageNet dataset to its variants that exhibit different domain shifts, such as style transfer and temporal variations in data collection.  The results indicate that CHBR consistently outperforms the baseline \textit{CLIP} across all datasets for all likelihood functions, validating the robustness of the concepts generated by LLMs to varying domain discrepancies.  However, compared to fine-grained classification tasks, the performance improvements achieved by CHBR on ImageNet variants are less significant when using Average Likelihood and Confidence Likelihood. It may be because ImageNet categories generally have more distinct visual differences, reducing the benefit of concepts acquired by CHBR, which focuses on both class descriptions and subtle class distinctions.  However, TTA Likelihood still yields notable improvements, with over a 4\% increase in average accuracy on OOD datasets, highlighting the flexibility of CHBR owing to its compatibility with test-time augmentation techniques and its effectiveness in handling more complex real-world scenarios. Therefore, we conclude that compared to Average Likelihood, which solely relies on prior knowledge derived from LLMs, Confidence Likelihood and TTA Likelihood can leverage feedback from CLIP or the complementary information of augmentation views to improve generalization. Furthermore, in contrast to TTA Likelihood, which requires gradient optimization, Confidence Likelihood operates entirely in a training-free manner. Additionally, Confidence Likelihood performs particularly well in fine-grained classification tasks, while TTA Likelihood is more robust in noisy scenarios where images may be affected by factors such as variations in image style or differing viewpoints.

\textbf{CHBR generalizes effectively across various visual encoder architectures with the same hyperparameters}. The generalization across various model architectures is crucial to CHBR, as it can eliminate the need for repeated hyperparameter tuning and concept sampling. To demonstrate this, we evaluate CHBR on five widely-used visual encoder architectures of CLIP, as shown in Fig. \ref{fig:3}. The results indicate that CHBR consistently improves performance compared to the baseline \textit{CLIP}, which employs the prompt template ``A photo of a \{class\}." for image classification across all architectures using identical hyperparameter settings. This improvement is particularly notable for Confidence Likelihood and TTA Likelihood, further substantiating the effectiveness of adaptive concept refinement tailored to specific inputs. Additionally, the performance gains achieved through CHBR tend to drop slightly as the capacity of the backbone models increases. It suggests that our concept-based guidance paradigm is particularly advantageous in more complex scenarios where the classification capacity of the backbone model is more constrained.
\subsection{Concept Analysis}
\label{conceptana}
\begin{figure*}[tb]
    \centering
        \centering
\includegraphics[width=\linewidth]{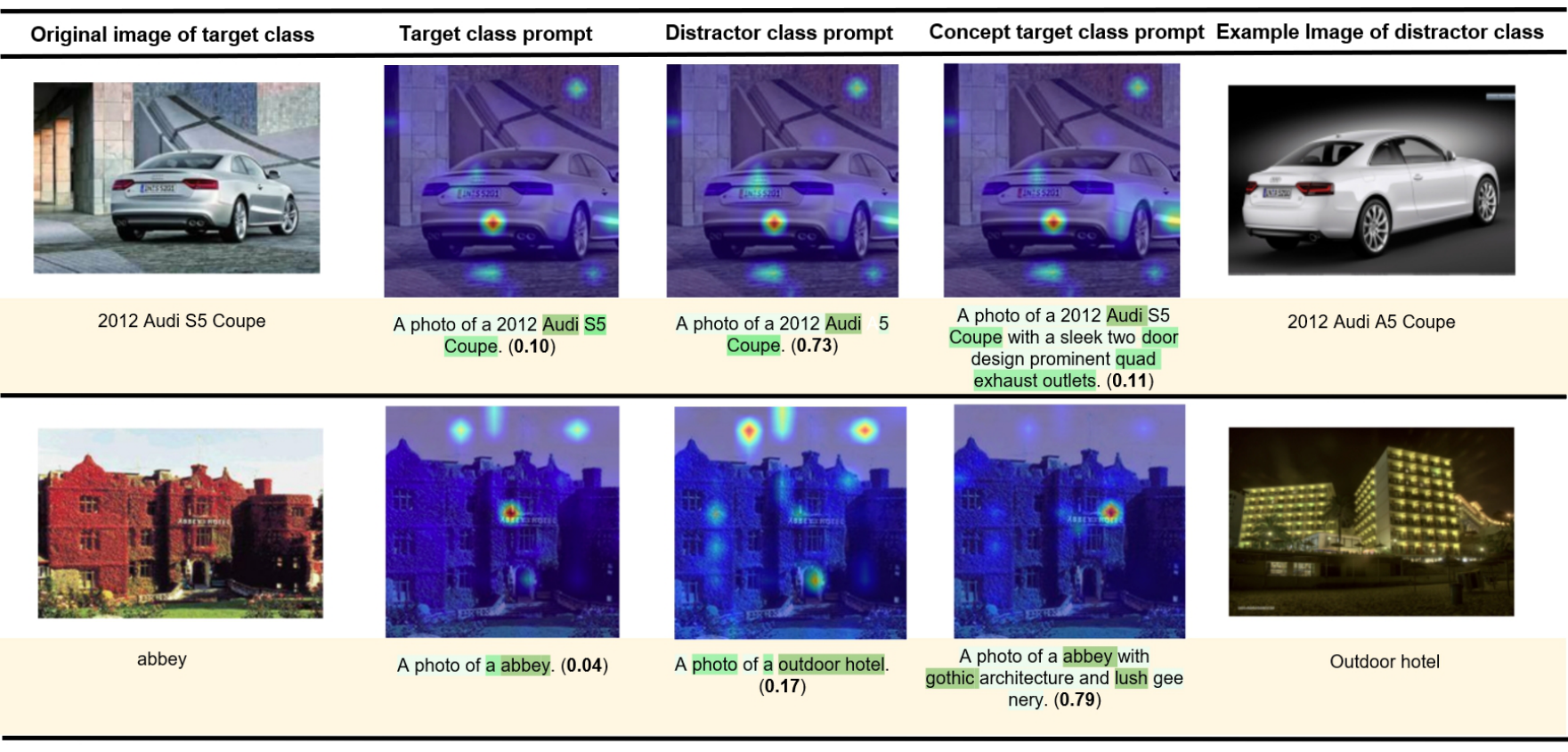}
    \caption{Visualization of the contribution of visual and text tokens to the final prediction for the target class prompt, distractor class prompt, and concept-based target class prompt. The distractor class refers to the set of classes into which the original image of the target class will be misclassified. In the visual heatmap, the higher intensity in red indicates a stronger influence on the prediction. For text tokens, darker shades of green correspond to higher contributions. Numerical values following the prompts are predicted probabilities.} 
    \label{fig:6}
\end{figure*}
\quad~\textbf{Discriminative concepts show advantages over descriptive concepts}.  CHBR posits that an efficient concept sampling space should not only encapsulate the characteristics of a class but also enhance the discriminability between the target class and other candidate classes. To empirically validate this hypothesis, we compare discriminative concepts generated by our importance sampler, augmented by a discriminative test, with descriptive concepts generated by prompting the same LLM to describe the class itself, following the previous approach ~\citep{pratt2023does}, as shown in Fig.~\ref{fig:4}. The results demonstrate that discriminative concepts outperform descriptive concepts across all datasets except Pets, thus empirically supporting the use of discriminative concepts to approximate the latent concept space. Furthermore, we emphasize that the performance of descriptive concepts is highly dependent on the inductive biases of LLMs. In cases where descriptive concepts are shared across classes, such as in the DTD (texture) dataset, the generated concepts may confound the classifier, leading to suboptimal performance.
\begin{table*}[h]
\centering
\caption{Representative concepts of each image class (Ranked by Probability)}
\label{concepttable}
\begin{adjustbox}{width=0.9\linewidth, center}
\begin{tabular}{cc>{\raggedright\arraybackslash}p{10cm}r}
\toprule
\textbf{Class} & & \textbf{Representative Concept} & \textbf{Probability} \\
\midrule
\multirow{5}{*}{2012 Audi A5 Coupe} 
 & 1 & A sleek coupe silhouette and distinctive single-frame grille & 0.3200 \\
 & 2 & A sleek, low-profile coupe design and distinctive LED headlights & 0.2400 \\
 & 3 & A sleek, aerodynamic silhouette and a prominent front grille & 0.2300 \\
 & 4 & A sleek, sporty coupe profile and distinctive Audi grille & 0.1400 \\
 & 5 & A streamlined, low-profile body and a distinctive Audi grille & 0.0700 \\
\midrule
\multirow{5}{*}{2012 Audi S5 Coupe} 
 & 1 & A sleek, low-profile coupe silhouette and distinctive front grille & 0.3300 \\
 & 2 & A sleek two door design prominent quad exhaust outlets & 0.2800 \\
 & 3 & A sleek coupe silhouette and aggressive sporty design & 0.1600 \\
 & 4 & A sleek, aerodynamic body and distinctive LED headlights & 0.1500 \\
 & 5 & A sporty coupe design and distinctive single-frame grille & 0.0800 \\
 \midrule
 \multirow{5}{*}{Abbey} 
 & 1 & Large stone arches and a tranquil garden setting & 0.2525 \\
 & 2 & Gothic architecture and lush greenery & 0.2222 \\
 & 3 & Large stained glass windows and a tranquil garden & 0.2020 \\
 & 4 & Tall stone arches and stained glass windows & 0.2020 \\
 & 5 & Stone architecture and lush green gardens & 0.1212 \\
\midrule
\multirow{5}{*}{Outdoor Hotel} 
 & 1 & A lush garden and outdoor pool area & 0.3000 \\
 & 2 & Inviting outdoor seating and landscaped gardens & 0.2800 \\
 & 3 & Landscaped gardens and architectural balconies & 0.1900 \\
 & 4 & A lush garden and outdoor seating area & 0.1900 \\
 & 5 & A spacious balcony overlooking a scenic landscape & 0.0400 \\
\bottomrule
\end{tabular}
\end{adjustbox}
\end{table*}

\textbf{Case Study}. To illustrate how concepts generated by LLMs contribute to rectifying misclassifications, we analyze the contribution of each image and text token to CLIP's classification outcomes. This is achieved by leveraging the attention maps from the final layer of CLIP's visual and textual encoders, where each attention head is weighted by gradients, following~\citet{chefer2021generic}, as shown in Fig.~\ref{fig:6}. Our analysis reveals two key insights: 1) The concept introduces external prior knowledge beyond the class label, aiding the discrimination between target and distractor classes. For example, CLIP assigns more attention to ``quad exhaust outlets", a distinguishing feature of the Audi S5, whereas the Audi A5 only has two outlets. 2) The concept guides CLIP to focus on semantically relevant regions of the image. For instance, when provided with concepts ``lush" and ``gothic", CLIP emphasizes the building in the image rather than the sky, which is incorrectly prioritized by class label-only prompts.  Additionally, to illustrate the concepts corresponding to each class in Fig.~\ref{fig:6}, we present only a selection of representative concepts in Table \ref{concepttable} due to space constraints. Specifically, we use CLIP’s text encoder to extract embeddings for each concept associated with a given class and employ a clustering algorithm to identify cluster centers. The concepts closest to these cluster centers are designated as representative concepts. Subsequently, other concepts are classified into different cluster centers, and the probability of each representative concept is computed based on the proportion of concepts assigned to the corresponding cluster center relative to the total number of sampled concepts.
It is important to note that this embedding-based measurement cannot capture subtle nuances among concepts. Consequently, the computed probabilities are not suitable for directly determining the importance weights of each concept in Eq.~(\ref{eq:weight}).

\subsection{Ablation Study}
\begin{figure}[htbp]
    \centering
        \centering
\includegraphics[width=\linewidth]{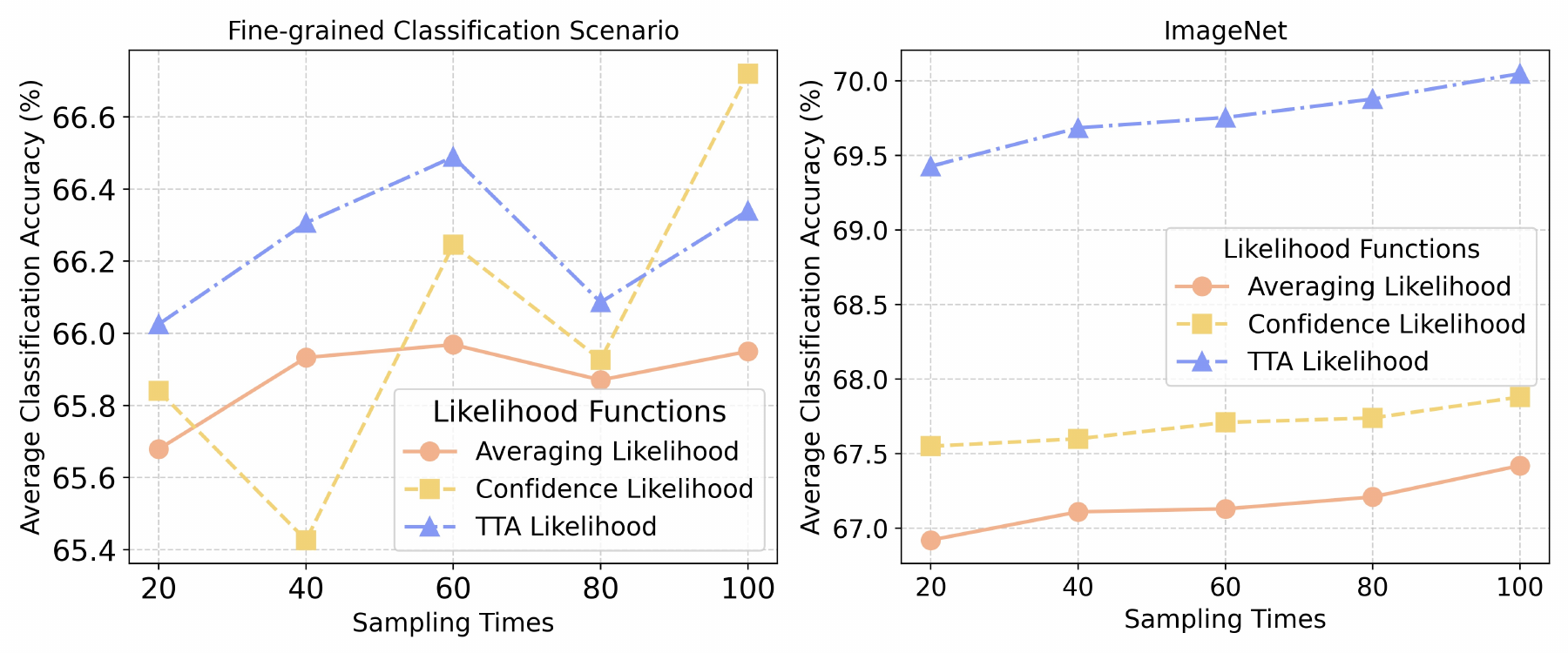}
    \caption{The accuracy trends of three likelihood functions on different image recognition scenarios as the number of sampling times increases. The left figure shows the accuracy on the ten fine-grained classification datasets, while the right figure presents the accuracy on the ImageNet dataset.} 
    \label{fig:5}
\end{figure}
\textbf{Effect of sampling times}. We examine the impact of different sampling times on the performance of CHBR in Fig. \ref{fig:5}.  As shown in Fig. \ref{fig:5} (Left), the results indicate that in the fine-grained classification scenario, the CHBR framework exhibits minimal performance variation across different sampling times for various likelihood functions. This stability can be attributed to the advantages of our sampling strategy, where the LLM is prompted to differentiate between multiple distractor classes rather than merely describe them, aligning with the findings in Sec. \ref{conceptana}. Furthermore, as depicted in Fig. \ref{fig:5} (Right), for ImageNet, which comprises 1000 categories, we observe a slight yet consistent improvement in performance as the number of sampling times increases. This suggests that when the label set is large and the inter-class differences are more pronounced, a higher sampling count enables a more precise approximation of the discriminative concept distribution $P(C_i|\mathcal{Y}_{/i}, Y_i)$, thereby yielding a closer approximation to the real-world concept distribution. Based on these observations, we set the number of sampling iterations to 100 in our main experiments to mitigate the need for hyperparameter tuning.
\begin{figure}[htbp]
    \centering
        \centering
\includegraphics[width=0.85\linewidth]{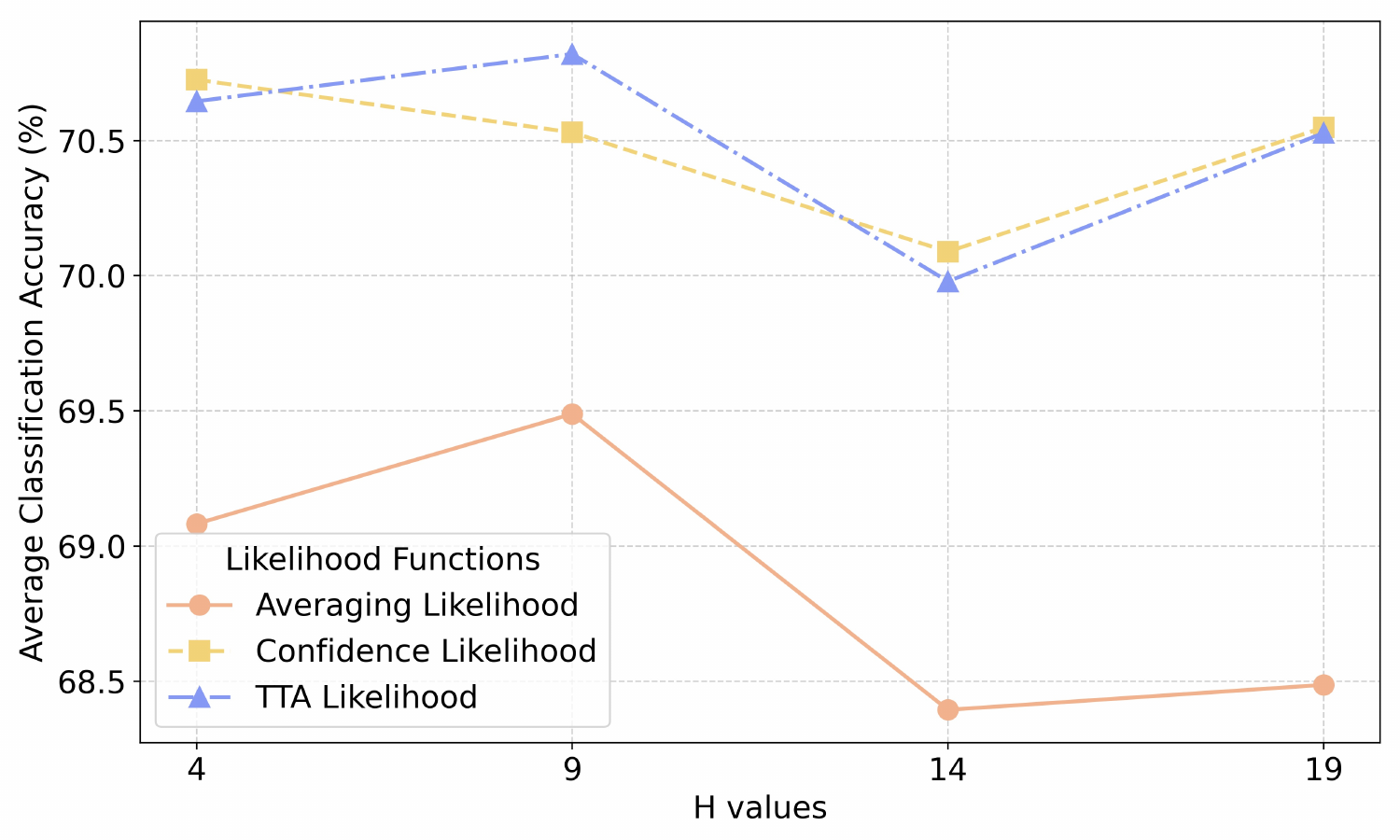}
    \caption{Comparison of average classification accuracy (\%) for different likelihood functions across various sizes of sampling windows $H$ on four datasets involving Pets, Flower102, DTD and UCF101.} 
    \label{fig:likelihood_comparison}
\end{figure}

\begin{table*}[bt]
\caption{Performance comparison of our proposed CHBR framework on two different models (GPT-4o mini and GPT-4 Turbo) across four datasets. We highlight the best and second-best results by bolding and underlining them, respectively. The sampling times for each class are set to 20 ($M_i=20$). Other hyperparameters are the same as in the main experiments.}
\centering
\label{tab:chbr_performance}
\begin{adjustbox}{width=0.8\textwidth, center}
\begin{tabular}{lcccccc}
\toprule
Model & Likelihood & Pets & Flower102 & DTD & UCF101 & Average \\
\midrule
\multirow{3}{*}{GPT-4o mini} 
& Averaging Likelihood & 88.67 & 68.48 & 50.18 & 69.00 & 69.08 \\
& Confidence Likelihood & 89.02 & 70.56 & 50.41 & 68.33 & 69.58\\
& TTA Likelihood & \textbf{89.51} & 70.52 & 51.65 & 69.13 & \underline{70.56} \\
\midrule
\multirow{3}{*}{GPT-4 Turbo} 
& Averaging Likelihood & 87.02 & 70.44 & 51.83 & 69.23 & 69.63 \\
& Confidence Likelihood & 88.76 & \underline{71.30} & \underline{52.01} & \underline{69.68} &70.44 \\
& TTA Likelihood & \underline{89.31} & \textbf{72.15} & \textbf{53.78} & \textbf{70.68} & \textbf{71.48} \\
\bottomrule
\end{tabular}
\end{adjustbox}
\end{table*}
\textbf{Effect of size of candidate set}.  As discussed in Sec.~\ref{sec:4.2}, an ideal proposal distribution for concept sampling should be conditioned on both the target class $Y_i$ and its complement $\mathcal{Y}_{/i}$ (i.e., all classes except $Y_i$), formally expressed as $q(C_i \mid \mathcal{Y}_{/i}, Y_i)$. In practice, to approximate this distribution, we randomly sample $H$ candidate classes from the overall image class set $\mathcal{Y}$ for each target class  $Y_i$. We then prompt LLMs to generate candidate concepts that effectively distinguish $Y_i$ from the selected $H$ candidate classes and the target class itself. However, this approach inherently involves a trade-off between sampling efficiency and the model’s discriminative capability. Intuitively, increasing $H$ allows for a more accurate approximation of the $q(C_i | \mathcal{Y}_{/i}, Y_i)$ while keeping the sampling time unchanged. Nevertheless, a larger $H$ also necessitates the LLM to differentiate among a broader set of candidate classes, thereby increasing the complexity of the reasoning process. To assess the impact of the sampling window size $H$, we evaluate the CHBR framework using different likelihood functions across varying values of $H$ on four datasets, as illustrated in Fig.~\ref{fig:likelihood_comparison}. The experimental results suggest that when employing the Averaging Likelihood, performance improves as $H$ increases from 4 to 9. However, beyond this point, further increases in $H$ lead to a decline in performance. Since Averaging Likelihood assumes that each concept contributes equally to the representation of the input image, it effectively reflects the quality of the sampled concept space. These findings indicate that while a larger $H$ enhances the quality of candidate concepts, an excessively large $H$ may exceed the model’s capacity, potentially leading to hallucination effects. Furthermore, in contrast to Averaging Likelihood, the other two likelihood functions exhibit greater robustness to noisy concepts, particularly when $H=19$. This observation further substantiates the effectiveness of our proposed framework, which integrates likelihood functions to dynamically adjust the weight of concepts for each input image, thereby enhancing both adaptability and reliability.

\textbf{Effect of size of various LLMs}. Table \ref{tab:chbr_performance} presents the performance comparison of our proposed CHBR framework on GPT-4o mini and GPT-4 Turbo, across four fine-grained classification datasets. GPT-4 Turbo may contain approximately 20 to 100 times more parameters than GPT-4o mini, as estimated based on their respective API pricing. The results indicate that GPT-4 Turbo generally outperforms GPT-4o Mini, with the highest average accuracy observed under the TTA Likelihood method. Notably, the performance gap between the models is more pronounced on datasets such as Flower102 and DTD.  These findings suggest that the increased model capacity enhances the quality of the concept generation process for approximating the concept prior in real-world scenarios, ultimately leading to improved classification accuracy.
\begin{table*}[bt]
\caption{Performance of MTA and the integration of MTA and the concepts generated by our proposed sampling algorithms ($\textrm{MTA}^\textrm{C}$). The subscript indicates the difference in performance between the two methods.}
\label{tab:zero_shot_mat}
\centering
\begin{adjustbox}{width=\linewidth, center}
\begin{tabular}{lccccccccccc}
\toprule
Method & SUN397 & Aircraft & EuroSAT & Cars & Food101 & Pets &  Flower102 & Caltech101 & DTD & UCF101 & Average
\\ \midrule 
MTA~\citep{matcvpr24}  & 64.98 & 25.32 & 38.71 & 68.05 & 84.95 & 88.22 & 68.26 & 94.13 & 45.59 & 68.11 & 64.63 \\
$\textrm{MTA}^\textrm{C}$  & 66.50$_{+1.52}$ & 27.12$_{+1.80}$ & 36.78$_{-1.93}$ & 68.36$_{+0.31}$ & 81.32$_{-3.63}$ & 89.97$_{+1.75}$ & 70.16$_{+1.90}$ & 92.13$_{-2.00}$ & 51.83$_{+6.24}$ & 68.75$_{+0.64}$ & 65.67$_{+0.61}$ \\
\bottomrule
\end{tabular}
\end{adjustbox}
\end{table*}

\begin{table*}[bt]
\caption{The performance comparison of our proposed CHBR framework using the original prompt and the efficient prompt across four datasets. We highlight the best and second-best results by bolding and underlining them, respectively. The hyperparameter is the same as in the main experiments.}
\centering
\label{tab:effieprompt}
\begin{adjustbox}{width=0.8\textwidth, center}
\begin{tabular}{lcccccc}
\toprule
Prompt Version & Likelihood & Pets & Flower102 & DTD & UCF101 & Average \\
\midrule
\multirow{3}{*}{Original Version} 
& Averaging Likelihood & 88.14 & 68.41 & 50.65 & 69.13 & 69.08 \\
& Confidence Likelihood & \textbf{90.08} & \textbf{71.30} & \underline{51.95} & 69.57 & \textbf{70.73} \\
& TTA Likelihood & \underline{89.89} & \underline{70.40} & \textbf{52.90} & 69.39 & \underline{70.65} \\
\midrule
\multirow{3}{*}{Efficient Version} 
& Averaging Likelihood & 89.62 & 69.31 & 49.41 & \underline{69.63} & 69.49 \\
& Confidence Likelihood & 89.29 & 69.79 & 49.82 & 69.23 & 69.53 \\
& TTA Likelihood & 89.86 & 69.63 & 51.30 & \textbf{70.13} & 70.23 \\
\bottomrule
\end{tabular}
\end{adjustbox}
\end{table*}

\section{Discussion}
\label{sec:discussion}
\textbf{Computation costs}. We evaluate the computational cost of CHBR from concept sampling and parameter optimization during the inference stage. Specifically, when the number of sampling iterations is set to $M$, and the verification count for each concept in the discriminative test is set to $Z$ with $K$ representing the number of classes in the dataset, the total number of queries to the LLM is $M \times Z \times K$.  If concepts are sampled sequentially via the chat API, this process can be time-consuming, taking approximately 15 hours for $K=100$. However, by utilizing the batch API\footnote{\url{https://platform.openai.com/docs/guides/batch/overview}}, where multiple queries can be processed within a single API call, results are guaranteed within 24 hours. It allows for sampling all concepts across all datasets in a single day. Additionally, the batch API is more cost-effective than submitting individual queries per call, with pricing at $0.075\$$ per 1 million input tokens and $0.300\$$ per 1 million output tokens for GPT-4o mini. Finally, the total experimental cost is around $130\$$. Regarding parameter optimization, both the Averaging Likelihood and Confidence Likelihood methods are training-free and do not require learnable parameters. In contrast, TTA Likelihood necessitates the optimization of a learnable parameter $\theta \in \mathbb{R}^K$, less computationally intensive than prompt-tuning-based zero-shot generalization methods such as TPT \citep{shu2022tpt}.

\noindent\textbf{Cost-efficient concept sampling}. To further minimize the number of queries to LLMs, given a specific class $Y_i$, we first extract the text embeddings of all candidate classes and identify the most similar class from the candidate set to serve as the distractor class set $L_i$. For each target-distractor class pair $(Y_i,L_{i,j})$, we perform ten sampling iterations during the decoding process of the LLMs to reduce the number of  queries. Subsequently, for all generated concepts associated with a given pair, we leverage the advanced discriminative capabilities of recent LLMs by conducting a discriminative test wherein the LLMs evaluate and differentiate these ten generated concepts in a single query. As a result, the total number of  queries  can be reduced to $M\times Z \times K/100$. The prompt used for this  query-efficient discriminative test is provided in Table \ref{prompt:discriminativetesteff} in the Appendix. Furthermore, we compare the performance of our proposed CHBR framework using the original prompt with that of the  query-efficient prompt in Table \ref{tab:effieprompt}. The results indicate that the efficient prompt leads to a slight decline in performance compared to the original prompt, demonstrating the inherent trade-off between efficiency and accuracy. However, despite this performance drop, our method utilizing the efficient prompt still outperforms the baseline models TPT and MTA, achieving approximately a 2\% improvement in accuracy across four datasets.

\noindent\textbf{Inference time}. We run all experiments on four NVIDIA 3090Ti GPUs. Since both Averaging Likelihood and Confidence Likelihood are training-free methods, their computational time is comparable to the baseline model, \textit{CLIP}, as the time required for computing the similarity between different concept-enhanced prompts and the image embedding is negligible on an NVIDIA 3090 Ti GPU  compared to the time for computing the similarity between class-specific prompt and the test image. Consequently, our primary focus is comparing TTA Likelihood with TPT and MTA because all three require additional training during inference. To ensure a fair comparison of execution times across methods, we report the inference time for these three approaches on the Caltech101 dataset using the same GPU. Our findings indicate that TPT, MTA, and TTA Likelihood take around 0.11 seconds, 0.01 seconds, and 0.04 seconds for each test image. For simplicity, we exclude model loading time and focus solely on the adaptation time during inference. 

\noindent\textbf{Integration with MTA}. While CHBR has demonstrated superiority over existing zero-shot generalization methods due to the incorporation of concepts, concept prior and the likelihood function, it is worth investigating whether these concepts could also enhance current zero-shot generalization approaches. TPT, for instance, necessitates the concatenation of soft prompts with standard prompts and subsequently optimizes the soft prompt embedding using classification loss, which incurs significant computational overhead for all concepts. Therefore, we primarily integrate MTA with the concept prior by refining raw descriptions (e.g., ``A photo of a {class}.") into conceptually enriched variants (e.g., ``A photo of a {class} with {concept}."). We compute the average embedding of all concept-enhanced prompts to represent the textual embedding for each class. The results, presented in Table \ref{tab:zero_shot_mat}, indicate that the concept prior enhances MTA's performance, demonstrating the potential of incorporating human-like prior knowledge into advanced VLMs. Since the combination of MTA and concepts underperforms on datasets such as EuroSAT, Food10, and Caltech101, it is essential to explore adaptive refinement of concept importance for various generalization methods and test images as a likelihood function. These explorations are reserved for future work.

\section{Conclusion}
\label{sec:conclusion}
Inspired by the human-like zero-shot image recognition process, we introduce a Concept-guided Human-like Bayesian Reasoning (CHBR) framework. CHBR models the concept used in human image recognition as latent variables and formulates this task by summing across potential concepts, weighted by a prior distribution and a likelihood. To address the trade-off between the efficiency of exploring the concept space and computational overhead, we enhance importance sampling with a discriminative test, prompting LLMs to generate concepts that aid in distinguishing candidate image classes. Additionally, we propose three likelihood functions to refine the importance of each concept, improving CHBR's adaptability to individual test images during inference. Specially, Averaging Likelihood and Confidence Likelihood are particularly effective in fine-grained image classification, while TTA likelihood demonstrates robustness across both ImageNet and fine-grained scenarios. Nevertheless,  the inference time of Averaging Likelihood and Confidence Likelihood is comparable to that of directly prompting CLIP for classification, achieving significantly lower latency than TTA Likelihood and other test-time adaptation baselines.

Beyond this work, future research could explore concept generation and refinement in a few-shot setting, where multiple labeled images per class are available, to align better the prior distribution elicited from LLMs with real-world applications. In addition, CHBR has the potential to be developed into a plug-and-play module that enhances raw descriptions (e.g.,``A photo of a \{class\}.") by using a conceptually enriched variant (e.g., ``A photo of a \{class\} with \{concept\}."). While we have preliminarily experimented with integrating the zero-shot method MTA~\citep{matcvpr24} with Averaging Likelihood, future work could also investigate how to adaptively refine the importance of concepts for various generalization methods and test images.  We leave these explorations for future work.
\backmatter

\section*{Declarations}

\textbf{Conflict of interest.} The authors have no conflict of interest to declare
that are relevant to the content of this article.

\begin{appendices}
\section{Details of concept sampling algorithm}
\label{app:1}
In Sec.\ref{sec:4.2}, we propose a large language model-driven sampler to implement Monte Carlo importance sampling from $q(C_{i}| \mathcal{Y}_{/i}, Y_i)$. Concretely,  for each class $Y_i$, we randomly draw $H$ candidate classes from the image class set $\mathcal{Y}$, forming a candidate set $L_{i,j}$ for $Y_i$. The LLM is then prompted to generate the candidate concept $C_{i,j}$ to describe $Y_i$ in a way that maximally distinguishes it from the other classes in  $L_{i,j}$. \textcolor{red}{The prompt for concept generation} is detailed in Table \ref{prompt:congen}. To assess the discriminability of $C_{i,j}$, we further adopt these $H$ distractor classes from $L_{i,j}$ to create a discriminative test, deciding whether LLM can accurately predict $Y_i$ from these distractor ones given $C_{i,j}$. \textcolor{red}{The prompt for the discriminative test} is provided in Table \ref{prompt:discriminativetest}. The success rate $s_{i,j}$ regarding that $C_{i,j}$ passing this test is recorded. 
Given that the true prior distribution $p(C_{i,j})$ is typically difficult to obtain, we approximate it using the success rate $s_{i,j}$. Finally, importance weight $w_{i,j}$ can be computed as:
\begin{small}
\begin{equation}
w_{i,j} = \frac{s_{i,j}}{q(C_{i,j}|\mathcal{Y}_{/i}, Y_i)},
\end{equation}
\end{small}where $q(C_{i,j}|\mathcal{Y}_{/i}, Y_i)$ equals to $1$ for simplification purpose following recent practice~\citep{ellis2024human}. The complete sampling algorithm is outlined in Algorithm 2.
\begin{table*}[ht]
    \centering
        \caption{Prompt for concept generation. The red text represents variables that will be replaced based on the specific datasets and image classes being tested. ``Core class" and ``Other Classes" will be instantiated by $Y_i$ and $L_{i,j}$, respectively.}
    \begin{tabular}{>{\raggedright\arraybackslash}p{0.95\textwidth}}
        \hline
        \textbf{SYSTEM PROMPT:} \\
        You are a visual concept proposer tasked with enhancing text descriptions for zero-shot image classification on the test dataset using CLIP.  \\
        Given: \\
        A core class from the test dataset. The set of other classes in the dataset.\\
        Task: \\
        Propose a concise, visually discriminative concept to append to the text description (i.e., ``A photo of \{Core Class\} with \{your concept\}") that helps CLIP better distinguish the core class from the other classes.

        Guidelines: 
        Analyze the unique visual characteristics of the core class compared to other classes.\\
        Propose a concept that captures these discriminative visual features. \\
        Ensure the concept is concrete, easily understandable by CLIP, and specific to the test dataset.\\

    Please remember the proposed concept should enable CLIP to classify images of the core class more accurately while minimizing confusion with other classes in the zero-shot setting.\\
        \hline
        \textbf{USER PROMPT:} \\ 
        Core class: \textcolor{red}{[Core class]}. Other classes: \textcolor{red}{[Other Classes]}.  Please remember to present the concept with ``The final concept is: " as a prefix in the last line.\\
        \hline
    \end{tabular}
    \label{prompt:congen}
\end{table*}
\begin{table*}[ht]
    \centering
    \caption{Prompt for discriminative test in our sampling strategy. The term ``Concept" will be replaced by $C_{i,j}$, and ``Class"  will be replaced by the concatenation of $Y_i$ and distractor classes.}
    \begin{tabular}{>{\raggedright\arraybackslash}p{0.95\textwidth}}
        \hline
        \textbf{SYSTEM PROMPT:} \\
        Please answer which class the concept belongs to. Just output the most possible class without external output.\\
        \hline
        \textbf{USER PROMPT:} \\ 
        Concept: \textcolor{red}{[Concept]}. Classes: \textcolor{red}{[Class]}.\\
        \hline
    \end{tabular}
    \label{prompt:discriminativetest}
\end{table*}

\begin{table*}[ht]
    \centering
    \caption{Prompt for efficient discriminative test in our sampling strategy, illustated in Sec. \ref{sec:discussion}. The term ``Concept" will be replaced by a list of $C_{i,j}$, and ``Class"  will be replaced by the concatenation of $Y_i$ and distractor classes.}
    \begin{tabular}{>{\raggedright\arraybackslash}p{0.95\textwidth}}
        \hline
        \textbf{SYSTEM PROMPT:} \\
        Please determine which class in image\_object\_classes the concept in the Python list image\_object\_concepts belongs to. Output a Python dict named predicted\_dict where key is the concept and value is the predicted class, wrapped with triple backticks ('''). \\
        \hline
        \textbf{USER PROMPT:} \\ 
        image\_object\_concepts= \textcolor{red}{[Concept]}. image\_object\_classes=: \textcolor{red}{[Class]}.\\
        \hline
    \end{tabular}
    \label{prompt:discriminativetesteff}
\end{table*}

\begin{algorithm}[ht]
\caption{LLM-driven Monte Carlo Importance Sampling Algorithm}
\label{alg:MCIS_LLM_Sampler}
\begin{algorithmic}
\REQUIRE Image class set $\mathcal{Y}$, number of candidate concepts $H$, number of repeats $Z$, size of sampling space $M_i$
\ENSURE Concept distribution $C_i$ and importance weights $\mathbf{w}_{i}$ for each class $Y_i$
\FOR{each class $Y_i$ in $\mathcal{Y}$}
    \STATE Initialize an empty set $\mathbf{w}_{i}$ for storing importance weights
    \FOR{$j = 1, 2, \dots, M_i$}
        \STATE Randomly sample $H$ candidate concepts from $\mathcal{Y}$ excluding $Y_i$, forming set $L_{i,j}$
        \STATE Query the LLM to generate candidate concept $C_{i,j}$ to describe $Y_i$ based on $L_{i,j}$
        \FOR{$z = 1, 2, \dots, Z$}
            \STATE Randomly sample $H$ distractor concepts from $\mathcal{Y}$, forming a distractor set
            \STATE Conduct a discriminative test to determine if the LLM can predict $Y_i$ from the distractor set given $C_{i,j}$
        \ENDFOR
        \STATE Record the success rate $s_{i,j}$ of $C_{i,j}$ passing the test over $Z$ runs
        \STATE Compute the importance weight $w_{i,j}$ as
        \[
        w_{i,j} = \frac{s_{i,j}}{q(C_{i,j}|\mathcal{Y}_{/i}, Y_i)} = s_{i,j}
        \]
        \STATE Save $w_{i,j}$ and concept $C_{i,j}$
    \ENDFOR
\ENDFOR
\RETURN The set of concepts $C_i$ and the corresponding importance weights $\mathbf{w}_{i}$ for each class $Y_i$
\end{algorithmic}
\end{algorithm}

\begin{table*}[t!]
\caption{Performance of zero-shot methods on ten fine-grained classification datasets with \textbf{error bar}.}
\label{tab:zero_shot_datasets_bar}
\centering
\begin{adjustbox}{width=\linewidth, center}
\begin{tabular}{lccccccccccc}
\toprule
Method & SUN397 & Aircraft & EuroSAT & Cars & Food101 & Pets &  Flower102 & Caltech101 & DTD & UCF101 & Average \\
\midrule
CHBR w/ Averaging   Likelihood & 65.54$_{\pm{0.02}}$ & 25.74$_{\pm{0.07}}$ & 48.96$_{\pm{0.34}}$ & 66.51$_{\pm{0.11}}$  & 82.46$_{\pm{ 0.02}}$ & 88.14$_{\pm{0.05}}$ & 68.41$_{\pm{ 0.24}}$ & 93.91$_{\pm{0.17}}$ & 50.65$_{\pm{0.29}}$ & 69.13$_{\pm{0.18}}$ & 65.95$_{\pm{0.02}}$ \\
CHBR w/  Confidence   Likelihood & 66.25$_{\pm{0.03}}$ & 26.28$_{\pm{0.29}}$ & 48.89$_{\pm{0.29}}$ & 66.62$_{\pm{0.04}}$ & 82.62$_{\pm{0.05}}$ & 90.08$_{\pm{0.27}}$  & 71.30$_{\pm{ 0.50}}$ & 93.59$_{\pm{0.14}}$ & 51.95$_{\pm{0.30}}$  & 69.57$_{\pm{0.11}}$  & 66.72$_{\pm{0.05}}$ \\
CHBR w/  TTA Likelihood  & 66.89$_{\pm{0.03}}$ & 28.20$_{\pm{0.49}}$ & 41.69$_{\pm{0.79}}$ & 68.86$_{\pm{0.03}}$ & 82.68$_{\pm{0.01}}$ & 89.89$_{\pm{0.10}}$ & 70.40$_{\pm{ 0.14}}$ & 92.74$_{\pm{0.16}}$ & 52.90$_{\pm{0.14}}$ & 69.39$_{\pm{0.31}}$ & 66.34$_{\pm{0.06}}$ \\
\bottomrule
\end{tabular}
\end{adjustbox}
\end{table*}

\end{appendices}

\clearpage
\bibliography{sn-bibliography}



\end{document}